\documentclass[10pt,twocolumn,letterpaper]{article}

\usepackage{iccv}              

\definecolor{iccvblue}{rgb}{0.21,0.49,0.74}
\usepackage[pagebackref,breaklinks,colorlinks,allcolors=iccvblue]{hyperref}
\usepackage{algorithm}
\usepackage{algorithmic}
\usepackage{flushend}
\usepackage{graphicx}
\usepackage{multirow}
\usepackage{bbding}
\usepackage{colortbl}
\usepackage{array}
\definecolor{best}{rgb}{1.0, 0.6, 0}
\definecolor{best2}{rgb}{1.0, 0.8, 0.6}

\def\paperID{9346} 
\def\confName{ICCV}
\def\confYear{2025}

\title{Frequency Autoregressive Image Generation with Continuous Tokens}

\author{Hu Yu$^{1}$
\;\;\;\;
Hao Luo$^{2}$
\;\;\;\;
Hangjie Yuan$^{2}$
\;\;\;\;
Yu Rong$^{2}$
\;\;\;\;
Jie Huang$^{1}$
\;\;\;\;
Feng Zhao$^{1}$
\;\;\;\; \\
$^1$ University of Science and Technology of China  $^2$ Alibaba Group, DAMO Academy \\
\url{https://yuhuustc.github.io//projects/FAR.html}
}

\begin{document}

\let\oldtwocolumn\twocolumn
\renewcommand\twocolumn[1][]{
    \oldtwocolumn[{#1}{
    \begin{center}
    \includegraphics[width=0.96\linewidth]{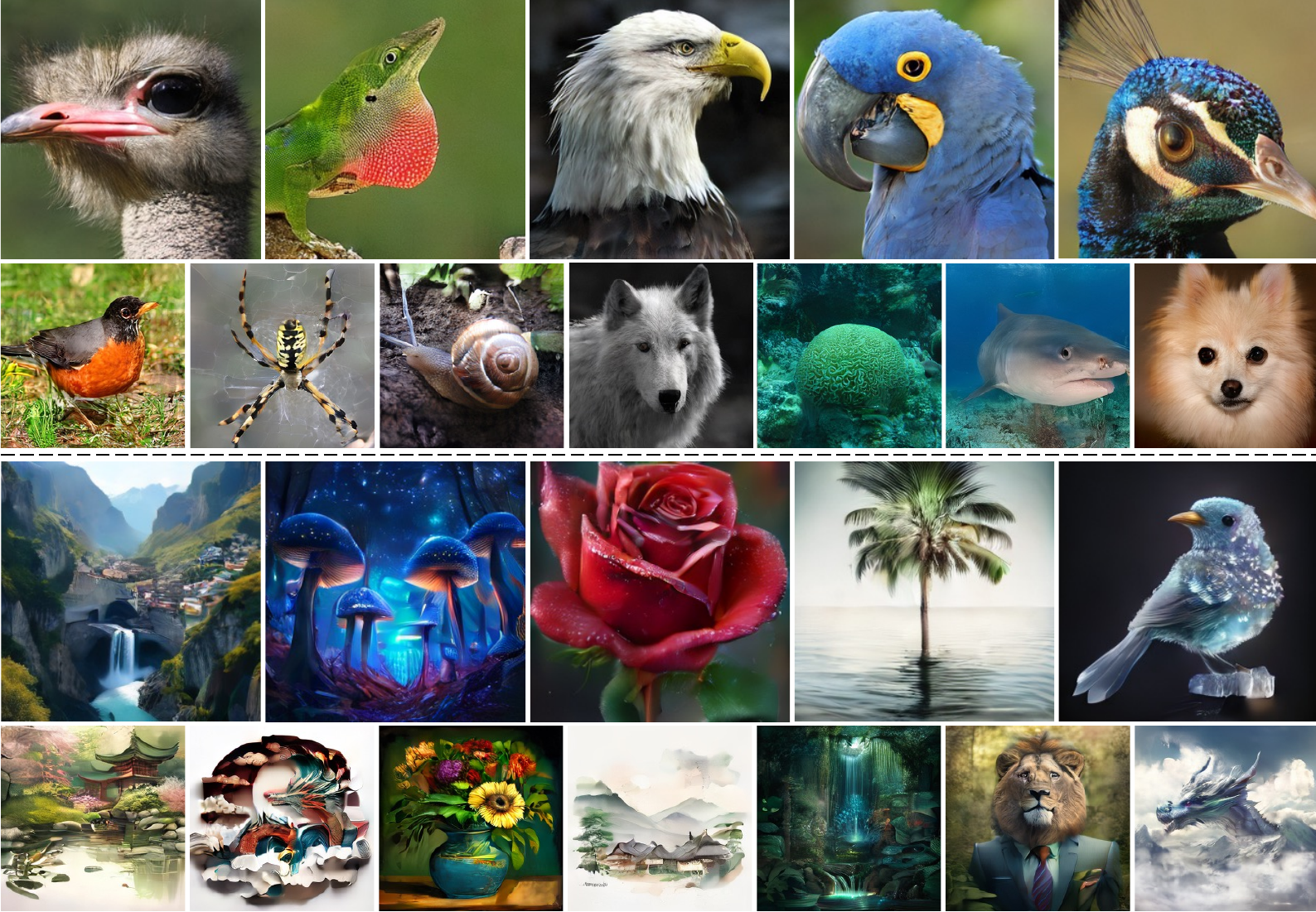}
    \setlength{\abovecaptionskip}{-0.05cm}
    \setlength{\belowcaptionskip}{-0.3cm}
    \captionof{figure}{\textbf{Samples from our FAR autoregressive model with continuous tokens.} Upper part: class-conditional generation. Lower part: text-to-image generation (with prompts in the supplementary material). All of these samples are generated in only 10 steps. }
    \label{fig:mainapp}
    \end{center}
    }]
}

\maketitle
\begin{abstract}
    Autoregressive (AR) models for image generation typically adopt a two-stage paradigm of vector quantization and raster-scan ``next-token prediction", inspired by its great success in language modeling. However, due to the huge modality gap, image autoregressive models may require a systematic reevaluation from two perspectives: tokenizer format and regression direction. 
    In this paper, we introduce the frequency progressive autoregressive (\textbf{FAR}) paradigm and instantiate FAR with the continuous tokenizer. 
    Specifically, we identify spectral dependency as the desirable regression direction for FAR, wherein higher-frequency components build upon the lower one to progressively construct a complete image. This design seamlessly fits the causality requirement for autoregressive models and preserves the unique spatial locality of image data.
    Besides, we delve into the integration of FAR and the continuous tokenizer, introducing a series of techniques to address optimization challenges and improve the efficiency of training and inference processes. 
    We demonstrate the efficacy of FAR through comprehensive experiments on the ImageNet dataset and verify its potential on text-to-image generation.

\end{abstract}

\section{Introduction}
\label{sec:intro}
Building upon autoregressive (AR) models, large language models (LLMs) \cite{devlin2018bert, raffel2020exploring, brown2020language, openai2022chatgpt} have unified and dominated language tasks with promising intelligence in generality and versatility, demonstrating a promising path toward artificial general intelligence (AGI).
To replicate the success of AR models in language processing and advance the unification of vision and language for AGI, pioneering researches \cite{esser2021taming, Dalle} have explored AR models for image generation. 
Resembling language processing, a typical AR paradigm for image generation involves two stages: 
\textbf{1)} Discretizing image data via vector quantization (VQ) with a finite, discrete vocabulary; 
and \textbf{2)} Flattening the quantized tokens into a 1-D sequence for next-token prediction. 
Based on this foundational paradigm, recent works \cite{chang2022maskgit, yu2023language, yu2024image, tian2024visual, han2024infinity, sun2024autoregressive, li2024autoregressive} have made inspiring advancements, demonstrating comparable, or even superior performance to diffusion models \cite{song2020score, rombach2022high} in image generation.

However, due to the huge modality gap between vision and language data, directly inheriting autoregressive generation from language to image is far from optimal. 
Text and image represent two distinct modalities: 
\textbf{1)} Text is discrete, causal/sequential, and arranged in 1-D; 
\textbf{2)} Image is continuous, non-causal/non-sequential, and arranged in 2-D. 
These differences introduce two crucial considerations for autoregressive image generation
\textbf{1)} \textit{Tokenizer format} (concrete vs. continuous); 
\textbf{2)} \textit{Regression direction} (incorporating image-specific causality). 
Regarding the tokenizer format, continuous tokenizer aligns more naturally with image data and induces less information loss~\cite{li2024autoregressive,fan2024fluid,rombach2022high} (more analyses are placed in the supplementary material). 
For the regression direction, raster \cite{sun2024autoregressive} or random \cite{li2024autoregressive} order fails to establish a causal sequence for images and can undermine inherent data priors, such as spatial locality.
Consequently, the current AR paradigm for image generation remains sub-optimal and necessitates further investigation.

In this paper, we rethink the autoregressive image generation paradigm from a \textit{spectral dependency} perspective. Spectral dependency represents the strong correlation between high-frequency image details and low-frequency structures, where higher-frequency components build upon lower-frequency foundations to progressively construct a complete image.
With this insight, we propose the frequency progressive autoregressive (FAR) paradigm. 
Specifically, FAR performs autoregressive generation across increasingly higher frequency levels, thereby inherently satisfying the causality requirements of AR models.
Within each frequency level, FAR simultaneously models the distributions of individual token and their inter-dependencies, effectively preserving the spatial locality of image data.

To instantiate FAR with continuous tokenizer, we identify the challenges of optimization difficulty and variance when modeling different frequency levels, and propose simplifying and re-weighting the diffusion modeling of these levels. Additionally, to enhance training and inference efficiency, we introduce the mask mechanism and frequency-aware diffusion sampling as techniques for applicable FAR deployment. 
These proposed techniques, coupled with the intrinsic harmony between spectral dependency and image data, endow FAR with a more compatible autoregressive paradigm for image generation.
Comprehensive experiments on the ImageNet dataset demonstrate the efficiency and scalability of FAR, wherein it significantly reduces inference steps to the number of frequency levels while maintaining high structural consistency.
We also extend FAR to text-to-image generation to further validate its potential. FAR achieves promising generation quality and high prompt alignment, utilizing much smaller model size, data scale, training compute, and inference steps, compared to previous text-to-image models.

Overall, our contributions can be summarized as follows:
\begin{itemize}
    \item We propose the FAR paradigm, leveraging the spectral dependency of image data. 
    FAR fits the causality requirement of AR models and preserves the spatial locality of image data, while being more sampling efficient.
    \item We delve into the instantiation of FAR with the continuous tokenizer, introducing a series of techniques to address the optimization challenges and improve the efficiency of both training and inference. 
    \item We demonstrate the effectiveness and scalability of FAR through comprehensive experiments on ImageNet dataset and further extend FAR to text-to-image generation.
\end{itemize}

\begin{figure*}[t]
    \begin{center}
	\includegraphics[width=0.97\linewidth]{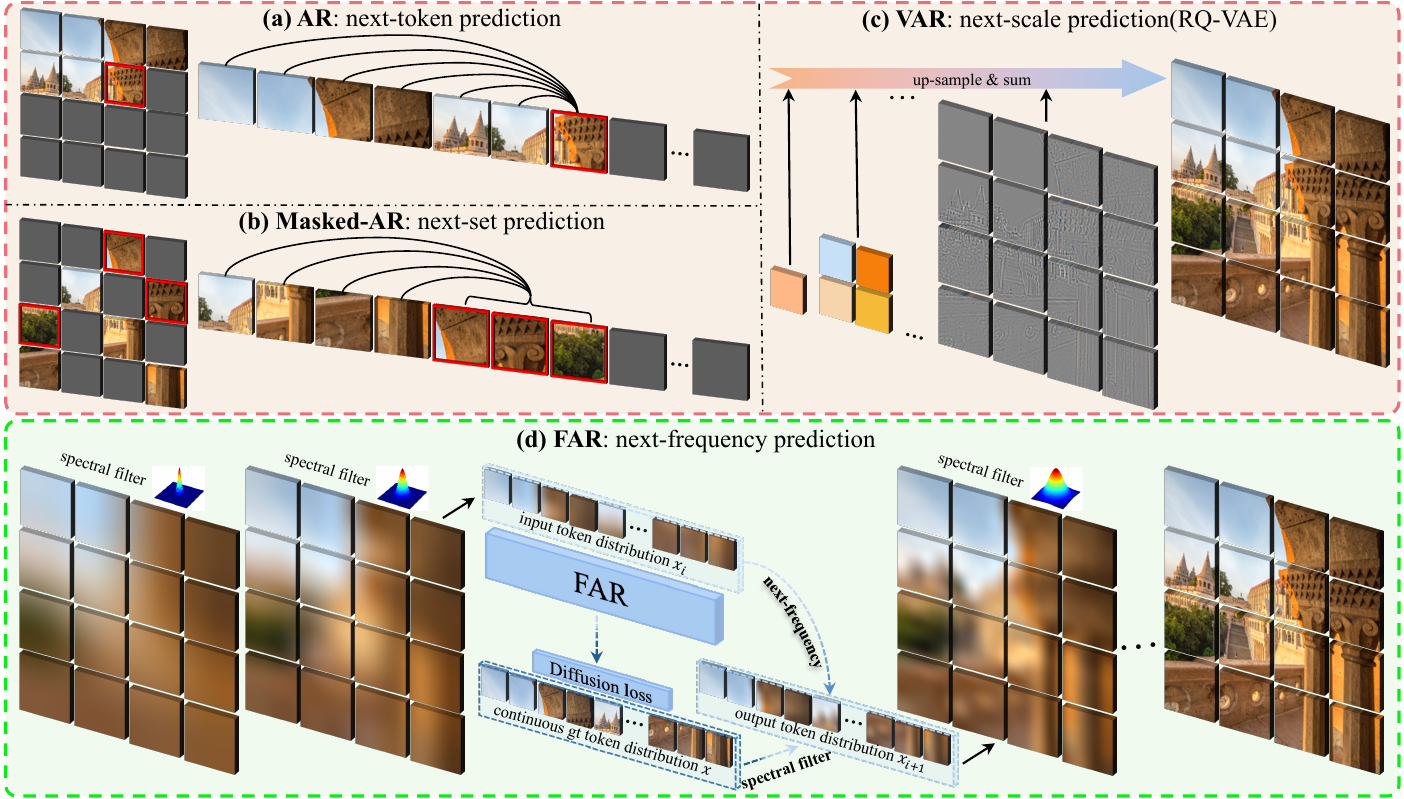}
    \end{center}
    \setlength{\abovecaptionskip}{-0.2cm}
    \setlength{\belowcaptionskip}{-0.2cm}
    \caption{\textbf{Regression direction paradigms in AR models for image generation}. (a) Vanilla AR: sequential next-token generation in a raster-scan order, from left to right, top to bottom; (b) Masked-AR: next-set prediction with random order, generating multiple tokens each step;
    (c) VAR: combines RQ-VAE and multi-scale, adding all scales to get the final prediction and necessitating customized multi-scale discrete tokenizer: (d) Ours FAR. We propose the next-frequency prediction paradigm leveraging the spectral dependency prior.}
    \label{fig:regression}
\end{figure*}

\section{Related Works}
\label{sec:related}

\subsection{Tokenizers in Autoregressive Models}
Most of existing AR models in the vision domain employ discrete tokens via vector quantization. The pioneering VQVAE \cite{van2017neural, razavi2019generating} proposes to quantize the latent space with a finite codebook, where each original token is replaced with the nearest discrete token in the codebook. VQGAN \cite{esser2021taming} further introduces adversarial training to VQVAE.
RQ-VAE \cite{lee2022autoregressive} proposes residual quantization (RQ). 
However, due to the significant information loss induced by quantization, the performance upper bound of AR methods may be limited. In contrast, continuous tokenizer \cite{rombach2022high}
aligns more naturally with image data and induces less information loss. Some recent works \cite{li2024autoregressive, fan2024fluid, deng2024autoregressive} also propose to employ a continuous tokenizer for autoregressive generation. In this paper, we also leverage continuous tokenizer and integrate it with our FAR paradigm. 
Comprehensive analyses of the tokenizers are available in the supplementary material.

\subsection{Autoregressive Models for Image Generation}
Autoregressive model is an important method for image generation, leveraging GPT-style \cite{radford2018improving} to predict the next token in a sequence. Raster-scan flattens the 2-D discrete tokens into 1-D sequences in a row-by-row manner. Most of previous image autoregressive models employ this manner, including VQGAN \cite{esser2021taming}, VQVAE-2 \cite{razavi2019generating}, Parti \cite{Parti}, DALL-E \cite{Dalle}, LlamaGen \cite{sun2024autoregressive}, etc. 

Besides this classical paradigm, some recent works make inspiring improvements over the raster-scan way. For example, MaskGIT \cite{chang2022maskgit} proposes masked-generation to generate next token set instead of next one token, substantially reducing the inference steps. 
MAR \cite{li2024autoregressive} proposes to combine continuous tokenizers with mask-based generation. 
Another type of methods adopt residual quantization. 
For example, RQ-VAE \cite{lee2022autoregressive} proposes modeling the residual with vector quantization. 
VAR \cite{tian2024visual} combines RQ-VAE with multi-scale, adding all scales to get the final prediction. This design enables the scale number to be the inference step. 
Direct compressing latent into 1-D sequence is also an interesting direction. 
TiTok \cite{yu2024image} compresses images into 1-D sequences with a modified autoencoder design. 
However, the extreme compression ratio may limit its performance. 

Some concurrent works \cite{pang2024randar, yu2024randomized, wang2024parallelized, pang2024next, ren2024flowar} propose other interesting AR methods. For example, RAR \cite{yu2024randomized} combines randomness and raster-scan and progresses from randomness to raster-scan for sequence generation. FlowAR \cite{ren2024flowar} combines the multi-scale design and flow model, and proposes multi-scale flow model for image generation.

Additionally, some multi-modal large models \cite{zhou2024transfusion, xie2024show, wang2024emu3} integrate the image generation ability into the AR models for unified understanding and generation. For instance, Show-O \cite{xie2024show} unifies autoregressive and (discrete) diffusion modeling to adaptively handle inputs and outputs of various and mixed modalities. Emu3 \cite{wang2024emu3} introduces a suite of multimodal models trained solely with next-token prediction.

Different from previous works, we propose FAR for autoregressive image generation with frequency progression, which fits the causality requirement of AR models and preserves the unique prior of image data.

\section{Preliminaries}
\label{sec:Preliminaries}

\subsection{Diffusion Loss for Continuous Tokens}
For the tokenizer in autoregressive models, the key is to model the per-token probability distribution, which can be measured by a loss function for training and a token sampler for inference. Following MAR \cite{li2024autoregressive}, we adopt diffusion models to solve these two bottlenecks for integrating continuous tokenizer into autoregressive models.

\noindent {\bf Loss function.} Given a continuous token $z$ produced by a autoregressive transformer model and its corresponding ground-truth token $x$, MAR employs diffusion model as loss function, with $z$ being the condition. 
\begin{equation}
    \mathcal{L}(z, x)=\mathbb{E}_{\varepsilon, t}\left[\left\|\varepsilon-\varepsilon_{\theta}\left(x_{t} \mid t, z\right)\right\|^{2}\right] .
\end{equation}
Here, $\varepsilon \sim \mathcal{N}(\mathbf{0}, \boldsymbol{I})$, and $x_{t}=\alpha_{t} x_{0}+\sigma_{t} \varepsilon$, with $\alpha_{t}$ being the noise schedule \cite{ho2020denoising}. The noise estimator $\varepsilon_{\theta}$, parameterized by $\theta$, is a small MLP network.

\noindent {\bf Token sampler.} The sampling procedure totally follows the inference process of diffusion model. Starting from $x_T \sim \mathcal{N}(\mathbf{0}, \boldsymbol{I})$, the reverse diffusion model iteratively remove the noise and produces $x_0 \sim p(x|z) $, under the condition $z$.

\subsection{Regression Direction}
Regression direction plays a crucial role in autoregressive models for image generation. In Figure \ref{fig:regression}, we illustrate the three prevailing regression direction paradigms for image autoregressive models. 

1) Vanilla AR (next-token prediction).  The ``next-token prediction" approach \cite{esser2021taming, sun2024autoregressive}, inspired by Large Language Models (LLMs) \cite{brown2020language}, employs a raster-scan method to flatten the interdependent 2-D latent tokens. This paradigm, however, violates the causal requirements of AR sequences. For example, the tokens at the front of the next row should depend on the tokens near it, instead of the token at the end of the last row. 
Additionally, this approach disrupts the spatial locality prior of images by predicting each token independently. Another limitation of vanilla AR is the inference speed, demanding the token length as step, which is unbearably slow for high-resolution image generation.

2) Masked-AR (next-set prediction). The mask-based generation method \cite{chang2022maskgit, li2024autoregressive}, derived from BERT \cite{devlin2018bert}, predicts the masked tokens given the unmasked ones. This paradigm enhances vanilla AR by incorporating randomness and predicting multiple tokens at every step. However, its generation quality at small sampling steps is restricted. More importantly, similar to the AR approach, Masked-AR violates the unidirectional dependency assumption of autoregressive models and neglects the image prior.

3) VAR (next-scale prediction). VAR \cite{tian2024visual} combines RQ-VAE \cite{lee2022autoregressive} with multi-scale, aggregating all scales to produce the final prediction. VAR maintains the spatial locality and adheres to the causality requirement. However, its multi-scale discrete residual-quantized tokenizer deviates from the commonly used tokenizers, necessitating specialized training. Furthermore, VAR requires training distinct tokenizers for various resolutions due to its multi-scale design. In contrast, our proposed method can seamlessly employ existing tokenizers, significantly enhancing its universality and scalability.
More importantly, we reveal that VAR paradigm demonstrates poor compatibility with the continuous tokenizer. 
Specifically, experiments combining VAR with the continuous tokenizer resulted in poor generation performance. Comprehensive results and analyses of the underlying reasons are provided in Section \ref{sec:experiments}.

\section{Methodology}
\label{sec:methodology}

\subsection{FAR}
\noindent {\bf Spectral dependency.} 
For the regression direction in autoregressive image generation, a pivotal challenge lies in harmonizing the causal sequence requirement with the inherent image prior. 
\textbf{In this paper, we identify spectral dependency as a distinctive image prior tailored to this context.} Specifically, images consist of low-frequency components that capture overall brightness, color, and shapes, alongside high-frequency components that convey edges, details, and textures \cite{russ2006image, wornell1996signal}. 
The generation of higher-frequency information intrinsically relies on the prior establishment of the lower one; for example, intricate details build upon and refine the foundational shapes and structure. 
This hierarchical process also mirrors human artistic painting, where an initial sketch outlines the overall structure, followed by the progressive addition of details.

Moreover, neural networks inherently exhibit similar spectral dependencies. DIP \cite{ulyanov2018deep} demonstrated that neural networks exhibit high impedance to high-frequency components while allowing low-frequency components to pass with low impedance. 
This indicates that neural networks naturally prioritize learning low-frequency information before progressing to the more complex high-frequency details. 
This characteristic aligns seamlessly with our design.

\noindent {\bf Autoregressive image generation with next-frequency prediction.} 
Leveraging the spectral dependency, we introduce the innovative next-frequency prediction for autoregressive image generation. For each image $x$, its intermediate input at frequency level $i \in \{1,2,..,F\}$ is formed as:
\begin{equation}
    x_i= \mathcal{F}^{-1} M_i \mathcal{F}x .
\end{equation}
Here, $F$ denotes the number of frequency levels, $\mathcal{F}$ and $\mathcal{F}^{-1}$ represent the Fourier transform and the inverse Fourier transform, respectively. $M_i$ denotes the spectral filter within level $i$. Higher frequency level $i$ represents more higher frequency information is preserved. 
For instance, $i=1$ retains only the lowest frequency component, capturing the average of the entire image, while $i=F$ retains all frequency components, representing the original image.
As shown in Figure \ref{fig:regression}(d), we decompose each image into various frequency levels and conduct autoregressive generation along these levels, progressively enhancing image clarity. 
This enables FAR to capture the interrelationships between these frequency components and learn the spectral mapping from lower to higher frequencies. 
Besides, within each frequency level, FAR employs bidirectional attention and predicts all tokens simultaneously at each step. 
This effectively models the dependencies between tokens in the 2-D plane, thereby preserving the spatial locality of images.
Furthermore, the inference step for generating an image with $n \times n$ tokens is also reduced from $\mathcal{O}\left(n^{2}\right)$ to linear complexity $\mathcal{O}\left(n\right)$.

\subsection{FAR with Continuous Tokens}
Besides the pioneering next-frequency prediction, we further delve into the combination of FAR and continuous tokens. 
We first identify and solve two primary challenges: optimization difficulty and variance in modeling token distributions across different frequency levels. 
Furthermore, we introduce the mask mechanism to enhance training efficiency and frequency-aware diffusion sampling strategy to accelerate inference.
More implementation details and flow chart visualization about training and inference processes are also available in the supplementary material.

\noindent {\bf Optimization difficulty: simplifying distribution modeling of diffusion loss.}
The diffusion loss in the continuous tokenizer models the distribution of per token. For FAR, diffusion loss needs to model $p\left(x_{i+1} \mid x_{i} \right)$ for $i \in [1, F-1]$, encompassing 
$F$ frequency levels. This multi-level distribution modeling is challenging for the relatively small MLP network. 
To mitigate this, we propose to directly model $p\left(x \mid x_{i} \right)$ for $i \in [1, F-1]$, and then filter $x$ to get $x_{i+1}$. 
This approach simplifies the optimization complexity by relaxing the diffusion loss to only model $x$.

\noindent {\bf Optimization variance: frequency-aware training loss strategy.}
Different frequency levels present varying optimization difficulties. 
Specifically, higher-frequency inputs are easier to predict, which can result in the optimization process being dominated by the more challenging low-frequency levels, thereby hindering the learning of higher-frequency details. 
To counteract this, we implement a frequency-aware training loss strategy that assigns higher loss weights to higher-frequency levels, ensuring balanced learning across all frequencies. Specifically, the loss weight is implemented in a sine curve as follows.
\begin{equation}
    w_i= 1 + \sin (\frac{\pi}{2} \times \frac{i}{F}),
\end{equation}
where $w_i$ is the loss weight of frequency level $i$.

\noindent {\bf Training efficiency: mask mechanism for reducing training cost and improving generation diversity.}
During the early steps of autoregressive generation, the network primarily needs to learn the low-frequency components, which are information-sparse. Consequently, utilizing all tokens from the preceding frequency level for subsequent predictions is redundant. 
To this end, we propose to incorporate the mask mechanism into FAR to leverage only a subset of tokens. 
Specifically, we devise a frequency-aware mask strategy that progressively increases the mask ratio for lower-frequency levels. The mask mechanism randomly masks $[r_i, 1]$ input tokens for the input tokens at frequency level $i$, where $r_i$ linearly transforms from $0.7$ to $0$.
This design effectively reduces the training cost and we find it also contributes to improving generation diversity.

\noindent {\bf Faster inference: frequency-aware diffusion sampling strategy.}
The frequency progression property of FAR also inspires us to employ fewer diffusion sampling steps for lower frequency levels, as diffusion models can sufficiently generate low-frequency information with small sampling steps.
Therefore, we devise the frequency-aware diffusion sampling step strategy that allocates progressively fewer steps to earlier frequency levels. 
Compared to the fixed diffusion sampling steps used in MAR, this strategy largely reduces the inference cost of the diffusion model.

\section{Experiments}
\label{sec:experiments}
In this section, we first introduce implementation details of our experiments in Section \ref{subsec:setup}. In Section \ref{subsec:classcond}, we present class-conditioned image generation results. In Section \ref{subsec:textcond}, we present  text-to-image image generation results.

\subsection{Setup} \label{subsec:setup}
\noindent {\bf Datasets.}
For class-conditioned generation, we adopt ImageNet \cite{deng2009imagenet} dataset. For text-to-image generation, we employ the JourneyDB \cite{sun2024journeydb} dataset with $\sim$4.19M image-text pairs and $\sim$3.57M internal data. By default, all images are center-cropped and resized to 256×256 resolution.

\noindent {\bf Training setup.}
We use the AdamW optimizer ($\beta_1 = 0.9$, $\beta_2 = 0.95$) \cite{loshchilov2017decoupled} with a weight decay of 0.02. Unless otherwise specified, for class condition, we train for 400 epochs with a batch size of 1024, and at an exponential moving average (EMA) rate of 0.9999.
For text condition, we train for 100 epochs with a batch size of 512 and EMA rate 0.99. 

\noindent {\bf Low-pass filters.}
We explore two frequency filtering types: (a) first down-sample then up-sample in the spatial domain, (b) low-pass filter in the Fourier domain. We find that they yield similar performance, as detailed in the supplement. By default, we empirically adopt type (a) for simplicity.

\noindent {\bf Models.}
We basically follow MAR \cite{li2024autoregressive} to construct our model, containing the AR transformer and diffusion MLP. 
The AR transformer has three model sizes: \textit{FAR-B} (172M), \textit{FAR-L} (406M), and \textit{FAR-H} (791M). The diffusion MLP is much smaller as shown in Table \ref{tab:MLP}. 
For text-to-image generation, we employ Qwen2-1.5B \cite{yang2024qwen2} as our text encoder and follow LI-DiT \cite{ma2024exploring} to reformat user prompts. Besides, we adopt cross attention for text condition injection.

\noindent {\bf Evaluation.}
We evaluate FAR on ImageNet with four main metrics, Fréchet inception distance (FID), inception score (IS), precision and recall, by generating 50k images. 
For text-to-image generation, we adopt MS-COCO and GenEval \cite{ghosh2024geneval} dataset. FID is computed over 30K randomly selected image-text pairs from the MS-COCO 2014 training set. The GenEval benchmark measures the alignment with the given prompt.

\begin{table*}[t]
    \setlength{\tabcolsep}{8pt}
    \centering
    \scriptsize
    \caption{\textbf{Performance comparisons on class-conditional ImageNet 256$\times$256 benchmark}.  ``$\downarrow$'' or ``$\uparrow$'' indicate lower or higher values are better. FAR achieves comparable generation quality in nearly all the evaluated metrics compared to the sota methods, with only 10 inference steps. The only exception of FID is attributed to the slighter lower diversity, which we find to greatly influence the FID metric. The \textit{Avg. Rank} in the last column represents the average ranking on the five indicators (including the additional inference steps) among the AR methods and our FAR-H. MAR (step=10) is evaluated using code and pretrained weights from their official GitHub repository.}
    \label{tab:Performance_ImageNet}
    \vspace{-5pt}
    \scriptsize
    \resizebox{0.92\textwidth}{!}{
    \begin{tabular}{l|l|cccc|cc|c}
    \toprule
    Tokenizer & Model & FID$\downarrow$ & IS$\uparrow$ & Precision$\uparrow$ & Recall$\uparrow$  & Params & Steps & Avg. Rank \\ 

    \midrule
    \rowcolor{lightgray} \multicolumn{9}{l}{\textbf{Diffusion Models}} \\
    \midrule

    \multirow{5}{*}{Continuous} 
    & ADM~\cite{dhariwal2021diffusion}  & 10.94 & 101.0 & 0.69 & 0.63  & 554M & 250 & -\\
    & LDM-4~\cite{rombach2022high} & 3.60  & 247.7  & 0.87  & 0.48  & 400M  & 250 & -\\
    & U-ViT-H/2-G~\cite{UViT} & 2.29  & 263.9  & 0.82  & 0.57  & 501M  & 50 & -\\
    & DiT-XL/2~\cite{peebles2023scalable} & 2.27 & 278.2 & 0.83 & 0.57 & 675M & 250 & -\\

    \midrule
    \rowcolor{lightgray} \multicolumn{9}{l}{\textbf{Autoregressive Models}} \\
    \midrule
    \multirow{9}{*}{Discrete} 
     & VQGAN~\cite{esser2021taming} & 15.78 & 74.3 &- &- & 1.4B & 256  & -  \\
     & ViT-VQGAN~\cite{vit-vqgan}  & 4.17  & 175.1  &- &- & 1.7B & 1024 & - \\
     & RQTran.~\cite{lee2022autoregressive} & 7.55  & 134.0 &- &- & 3.8B & 68 & -\\
     & MaskGIT~\cite{chang2022maskgit}  & 6.18  & 182.1  & 0.80 & 0.51 & 227M & 8 & 8 \\
     & LlamaGen \cite{sun2024autoregressive} & 2.81 & 311.6 & 0.84 & 0.54 & 3.1B & 576 & 5\\
     & VAR \cite{tian2024visual} & 3.30 & 274.4 & 0.84 & 0.51 & 310M & 10 & 2\\
     & PAR \cite{wang2024parallelized} & 2.88 & 262.5 & 0.82 & 0.56  & 3.1B & 51 & 6\\
     & RandAR \cite{pang2024randar} & 2.55 & 288.8 & 0.81 & 0.58  & 343M & 88 & 1\\
     & Open-MAGVIT2 \cite{luo2024open} & 3.08 & 258.3 & 0.85 & 0.51  & 343M & 256 & 4\\
    \midrule
    \multirow{5}{*}{Continuous}
    & MAR-H \cite{li2024autoregressive} & 1.55 & 303.7 & 0.81 & 0.62 & 943M & 256 & 2\\
    & MAR-H \cite{li2024autoregressive} & 9.32 & 207.4 & 0.71 & 0.47 & 943M & 10  & 9 \\
    \cline{2-9} 
    & \textbf{FAR-B} & 4.26 & 248.9 & 0.79 & 0.51 & 208M & 10 & 7\\
    & \textbf{FAR-L} & 3.45 & 282.2 & 0.80 & 0.54 & 427M & 10 & 3\\
    & \textbf{FAR-H} & 3.21 & 300.6 & 0.81 & 0.55 & 812M & 10 & 2\\
    \bottomrule
    \end{tabular}
    }
\end{table*}

\begin{table}[t]
    \setlength{\tabcolsep}{10pt}
    \centering
    \caption{\textbf{Scaling of denoising MLP in Diffusion Loss}. The denoising MLP is small and efficient, modeling the per-token distribution. Settings: FAR-L, 400 epochs, ImageNet 256x256.}
    \label{tab:MLP}
    \vspace{-3mm}
    \begin{tabular}{crr|cc}
    \toprule
    \multicolumn{3}{c|}{\textit{MLP}}  &  \multicolumn{2}{c}{\textit{Metrics}}  \\
    \midrule[0.1em]
    Depth & Width & \#Params  & FID$\downarrow$ & IS$\uparrow$ \\ 
    \midrule
     3 & 256  & 2M   & 3.83 & 278.2    \\
     3 & 512  & 6M   & 3.66 & 280.0    \\
     3 & 1024 & 21M  & 3.45 & 282.2    \\
     3 & 1536 & 45M  & 3.38 & 284.9    \\
    \bottomrule
    \end{tabular}
    \vspace{-3mm}
\end{table}

\subsection{Class-conditional Image Generation}  \label{subsec:classcond}

\noindent {\bf Scaling of the autoregressive transformers and denoising MLP.}
We investigate the scaling of both the autoregressive transformer and the diffusion loss model in Table \ref{tab:Performance_ImageNet} and Table \ref{tab:MLP}. The autoregressive transformer takes the main burden of modeling the frequency dependency and mapping, thus also accounting for the majority of the parameters. We find that the size of FAR transformer significantly affects the performance. When scaling up the FAR transformer, the performance is consistently improved.

For the denoising MLP, the requirement to model only the per-token distribution, combined with our distribution modeling simplification strategy, allows a small MLP (e.g., 2M) to achieve competitive performance. As expected, increasing the MLP width helps improve generation quality.

\begin{table*}[t]
    \setlength{\tabcolsep}{4pt}
    \centering
    \caption{\textbf{Performance comparisons on text-to-image task.}. Metrics include MS-COCO zero-shot FID-30K and GenEval benchmark. Please note that FAR employs much smaller model size, training data, GPU costs, and inference steps. We do not intend to demonstrate that FAR achieves cutting-edge performance, but rather to verify its potential in achieving high efficiency and promising results.
     }
    \label{tab:Performance_T2I}
    \vspace{-5pt}
    \resizebox{1.01\textwidth}{!}{
    \begin{tabular}{l | l | c | c c c c c c | c| c c c c}
    \toprule
    \multirow{2}{*}{Tokenizer} & \multirow{2}{*}{Model} & MS-COCO & \multicolumn{7}{c|}{GenEval} &  \multirow{2}{*}{Params} & Training & A100 & Infer \\
    \cline{4-10}
    & & {FID-30K$\downarrow$} & Sing-O. & Two-O. & Count. & Color & Pos. & Color-A. & Overall  & & Data & Days & Steps\\
    
    \midrule
    \rowcolor{lightgray} \multicolumn{14}{l}{\textbf{Diffusion Models}} \\
    \midrule
    
    \multirow{3}{*}{Continuous} & LDM \cite{rombach2022high}  & \hspace{-4pt}12.64 & 0.92 & 0.29 & 0.23 & 0.70 & 0.02 & 0.05 & 0.37 & 1.4B & - & 6250 & 250 \\
    & DALL-E 2 \cite{DALLE2}  & \hspace{-4pt}10.39 & 0.94 & 0.66 & 0.49 & 0.77 & 0.10 & 0.19 & 0.52 & 4.2B & 650M & - & 250\\
    & SD3 \cite{SD3}  & - & 0.98 & 0.84 & 0.66 & 0.74 & 0.40 & 0.43 & 0.68 & 8B & - & - & 50\\
    
    \midrule
    \rowcolor{lightgray} \multicolumn{14}{l}{\textbf{Autoregressive Models}} \\
    \midrule
    
    \multirow{5}{*}{Discrete}
    & DALL-E \cite{Dalle}  & 27.50 & - & - & - & - & - & - & -  & 12B & 250M & - & 256 \\
    & CogView2 \cite{Cogview2}  & 17.50 & - & - & - & - & - & - & -  & 6B & 35M & - & - \\
    & Muse \cite{Muse}  & 7.88 & - & - & - & - & - & - & - & 3B & 460M & 2688 & - \\
    & Parti \cite{Parti}  & 7.23 & - & - & - & - & - & - & - & 20B & 4.8B & - & 256  \\
    & LlamaGen \cite{sun2024autoregressive} & - & 0.71 &0.34 &0.21 &0.58 &0.07 &0.04 &0.32 & 775M & 60M & - & 256 \\
    \midrule
    \multirow{1}{*}{Continuous} & \textbf{FAR}  & 13.91 & 0.85 & 0.29 & 0.31 & 0.59 & 0.06 & 0.09 & 0.37 & 564M & 7.8M & 24 & 10 \\
    \bottomrule
    \end{tabular}
    }
\end{table*}

\begin{figure}[h]
    \begin{center}
	\includegraphics[width=0.98\linewidth]{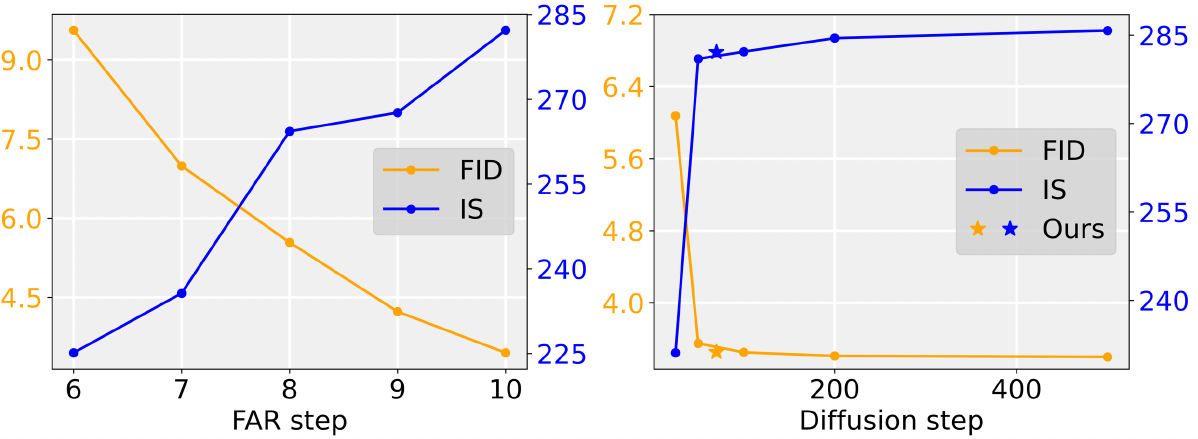}
    \end{center}
    \setlength{\abovecaptionskip}{-0.2cm}
    \setlength{\belowcaptionskip}{-0.2cm}
    \caption{\textbf{Sampling steps of FAR and diffusion loss}. }
    \label{fig:step}
    \vspace{-3mm}
\end{figure}

\noindent {\bf Sampling steps of FAR and diffusion loss.}
The training of the FAR adopts the maximum of $F$ autoregressive step. 
For the inference, however, we can flexibly change the autoregression step and adopt fewer steps than $F$. 
Specifically, given $x_i$, FAR directly model $x$. In the next autoregressive step, we can filter $x$ to get a flexible next frequency level, i.e. $x_{i+2}$ instead of $x_{i+1}$, enabling the dynamic autoregression steps of FAR. Figure \ref{fig:step} depicts the generation performance under different FAR autoregression steps, where a higher step consistently achieves better performance. 

The training of the denoising MLP employs a 1000-step noise schedule following DDPM \cite{ho2020denoising}. 
During inference, MAR verifies that fewer sampling steps ($T=100$) are sufficient for generation. 
We further demonstrate that our frequency-aware diffusion sampling achieves comparable results with fewer steps. 
Specifically, we linearly shift the sampling steps for $T=40$ to $T=100$, achieving an average sampling step of $T=70$. 
This saves $30\%$ inference time of the diffusion model. 
Figure \ref{fig:step} shows that our sampling strategy achieves comparable results with fewer steps.

\begin{table}[ht]
    \setlength{\tabcolsep}{16pt}
    \centering
    \caption{\textbf{Combining VAR and the continuous tokenizer}.}
    \vspace{-3mm}
    \label{tab:VAR}
    \begin{tabular}{cc|cc}
    \toprule
    \multicolumn{2}{c|}{\textit{VAR components}}  &  \multicolumn{2}{c}{\textit{Metrics}}  \\
    \midrule[0.1em]
    RQ & Multi-scale  & FID$\downarrow$ & IS$\uparrow$ \\ 
    \midrule
     \CheckmarkBold & \CheckmarkBold   & 75.35  & 33.2  \\ 
     \XSolidBrush   & \CheckmarkBold   & 33.57  & 96.8  \\ 
    \bottomrule
    \end{tabular}
    \vspace{-5mm}
\end{table}

\noindent {\bf The compatibility of VAR and diffusion loss.}
As we have noted in Section \ref{sec:Preliminaries}, the VAR paradigm demonstrates poor compatibility with the continuous tokenizer. 
The reasons are mainly two-fold. 
1) The RQ manner is highly accuracy-sensitive for the prediction at every step for the continuous tokenizer. The RQ in VAR up-samples the prediction at each scale to the full latent scale and adds them all to get the final output. This requires highly accurate predictions at each scale. However, the exposure bias problem \cite{bengio2015scheduled, arora2022exposure} of AR models induces inevitable error accumulation, deviating from the above requirement. 
2) The per-token distribution modeling task in the VAR paradigm is prohibitively challenging for the diffusion loss. The tokens in different scales differ significantly in both the numeric range and receptive field. 
As shown in Table \ref{tab:VAR}, directly combining VAR (RQ + mulit-scale) with the continuous tokenizer yields poor performance. 
Besides, we also try to remove the RQ design. 
The performance is improved due to no error accumulation in the residual paradigm. 
However, the performance still lags behind sota due to the gap in multi-scale. The visual results are available in the supplementary material.

Note that FlowAR \cite{ren2024flowar} combines VAR with continuous tokenizer, via directly modeling the distribution of the whole image with multi-scale diffusion model (flow model), instead of the token-wise distribution. It is thus more like a multi-scale diffusion model than a AR model.

\noindent {\bf Main results.}
In Table \ref{tab:Performance_ImageNet}, we list the comprehensive performance comparison with previous methods. We explore various model sizes and train for 400 epochs. 
Compared to most of the AR methods, our FAR is more efficient requiring fewer inference steps. 
Our method is superior to the VQGAN series with much smaller model size and inference steps. For recent works, like VAR and MAR, our method is also comparable in visual quality (indicated by IS and Perception metrics). 
Note that the lag in the FID metric is attributed to the slightly lower diversity (indicated by the Recall metric), which we find the FID metric is very sensitive to. 
Figure \ref{tab:Performance_ImageNet} shows qualitative results. We leave more visual results on ImageNet in the supplementary material.

\begin{figure*}[t]
    \begin{center}
	\includegraphics[width=0.95\linewidth]{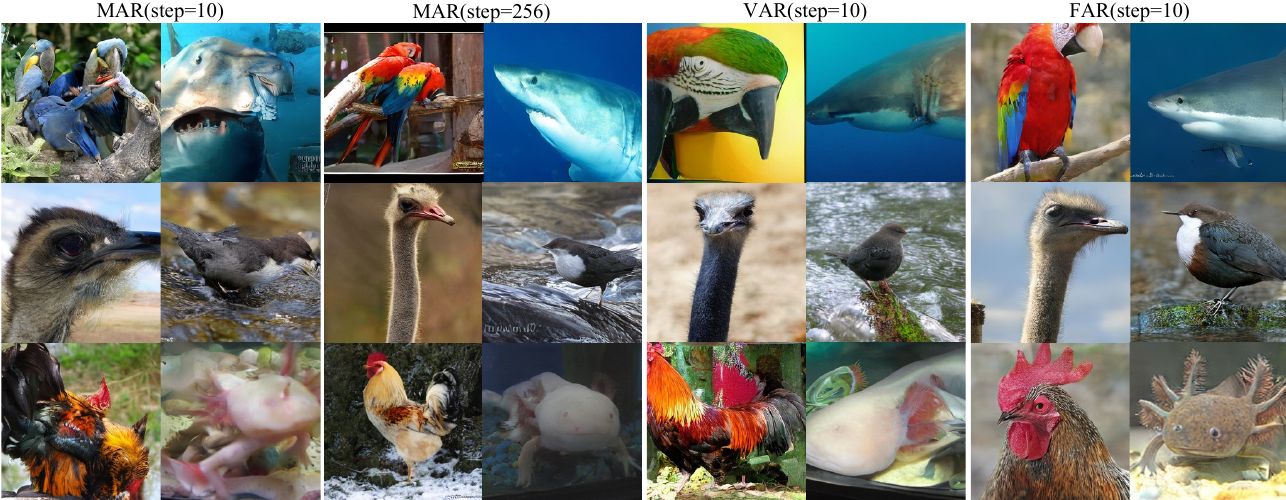}
    \end{center}
    \setlength{\abovecaptionskip}{-0.2cm}
    \setlength{\belowcaptionskip}{-0.2cm}
    \caption{\textbf{Visual comparisons with the representative MAR and VAR methods with 10 inference steps}. Thanks to the intrinsic harmony with image data, our FAR can generate high-quality images with consistent structures and fine details with only 10 steps.}
    \label{fig:comparison}
\end{figure*}

\noindent {\bf Visual comparison with MAR and VAR.} As shown in Figure \ref{fig:comparison}, the mask mechanism in MAR induces poor architecture under small inference steps. 
The discrete tokenizer in VAR may also limit the performance limit and has difficulty in generating images with complex composition. 
In contrast, due to the intrinsic harmony with image data, FAR can generate high-quality images with consistent structures and fine details with only 10 steps.

\begin{table}[t]
    \setlength{\tabcolsep}{6pt}
    \centering
    \caption{\textbf{Ablations on the effectiveness of the proposed techniques}. Specific meanings of the abbreviations are in the ablation part. Settings: FAR-L, MLP size 21M, 400 epochs.}
    \vspace{-3mm}
    \label{tab:Ablation}
    \begin{tabular}{ccc|cccc}
    \toprule
    \multicolumn{3}{c|}{\textit{Ablations}}  &  \multicolumn{4}{c}{\textit{Metrics}}  \\
    \midrule[0.1em]
    DMS & Mask & FTL  & FID$\downarrow$ & IS$\uparrow$ & Pre$\uparrow$ & Rec$\uparrow$ \\ 
    \midrule
    \CheckmarkBold & \XSolidBrush    & \XSolidBrush    & 13.47 & 281.4 & 0.89 & 0.11   \\
    \CheckmarkBold & \CheckmarkBold  & \XSolidBrush    & 4.11 & 288.9 & 0.79 & 0.51   \\ 
    \CheckmarkBold & \CheckmarkBold  & \CheckmarkBold  & 4.05 & 290.2 & 0.80 & 0.52   \\
    \bottomrule
    \end{tabular}
    \vspace{-5mm}
\end{table}

\begin{figure*}[t]
    \begin{center}
	\includegraphics[width=0.95\linewidth]{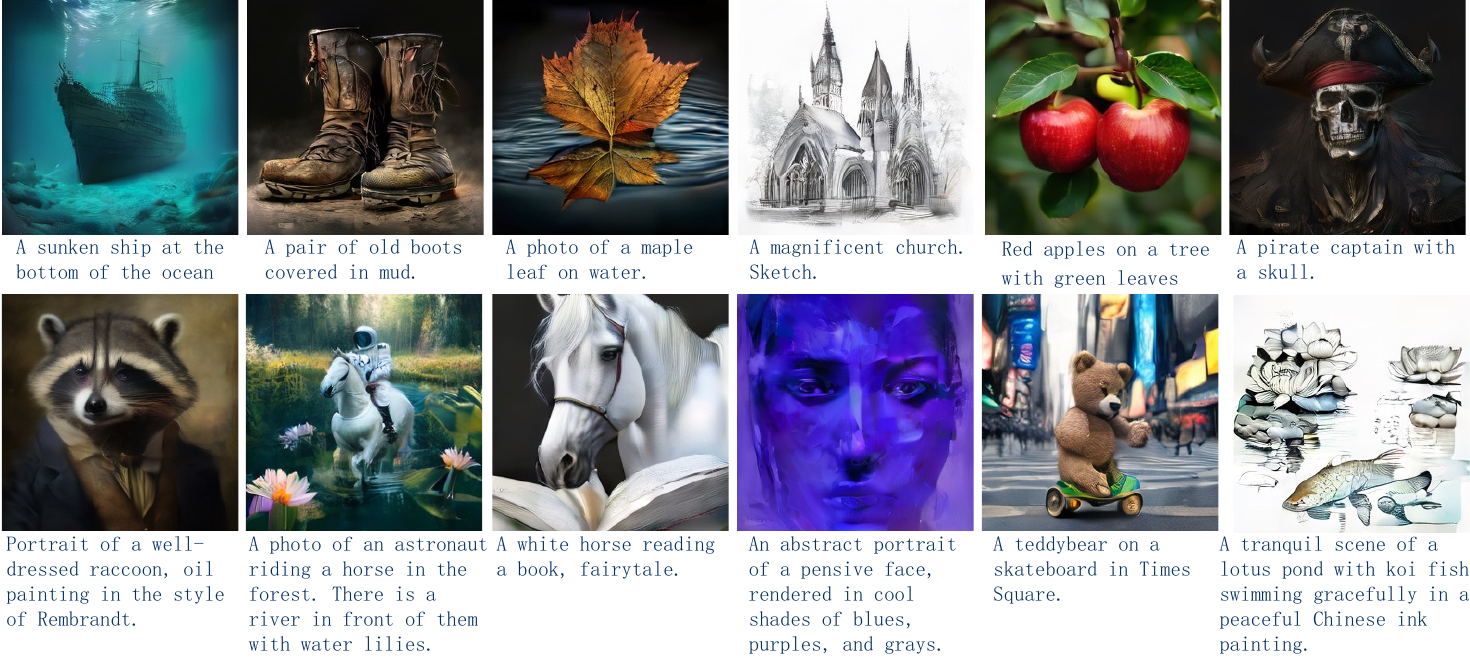}
    \end{center}
    \setlength{\abovecaptionskip}{-0.2cm}
    \setlength{\belowcaptionskip}{-0.2cm}
    \caption{\textbf{More visual results of the text-to-image autoregressive generation at 256x256 resolution}. }
    \label{fig:T2I}
\end{figure*}

\noindent {\bf More ablations.}
We also conduct extensive ablations to verify the effectiveness of our method, including:
\textbf{S1}) \textbf{DMS}: Diffusion loss Distribution Modeling Simplification strategy. 
\textbf{S2}) \textbf{Mask} mechanism. The mask mechanism improves the training efficiency by about $50\%$. 
\textbf{S3}) \textbf{FTL}: Frequency-aware Training Loss strategy. 
As shown in Table \ref{tab:Ablation}, with technique S1, FAR can already generate high-quality images (indicated by IS and Perception). The low FID is attributed to its lower diversity (indicated by Recall). 
With technique S2, the randomness in the mask mechanism compensates for the diversity and thus achieves significant FID improvement. 
We depict visual comparisons of this ablation in the supplementary material.

\subsection{Text-to-Image Generation}  \label{subsec:textcond}

\noindent {\bf Main results.}
In Table \ref{tab:Performance_T2I}, we depict the performance comparison on text-to-image generation task. Previous methods in this task usually employ substantially large model parameters, web-scale datasets, and unbearable computation costs. FAR can beat the classical DALL-E, CogView2 and LlamaGen, and achieve comparable performance to the recent sotas, with significantly smaller training and inference costs. The concurrent works \cite{fan2024fluid, deng2024autoregressive} also verify the effectiveness of continuous tokens on text-to-image generation with substantially more training resources. 
In Figure \ref{tab:Performance_ImageNet} and Figure \ref{fig:T2I}, we show the visual results on text-to-image generation. FAR can generate high-quality images with coherent structures and complex composition in 10 steps. 

\section{Conclusion}
\label{sec:conclusion}
In this paper, we propose the FAR paradigm and instantiate FAR with the continuous tokenizer. 
Specifically, we identify spectral dependency as the desirable regression direction for FAR. Besides, we delve into the integration of FAR and the continuous tokenizer. 
We demonstrate the efficacy and scalability of FAR through comprehensive experiments on class-conditional generation and further verify its potential on text-to-image generation.

\newpage
{
    \small
    \bibliographystyle{ieeenat_fullname}
    \bibliography{main}
}


\newpage
\appendix

\begin{center}
    \Large{\textbf{Appendix}}
\end{center}

This supplementary document is organized as follows: 

Section \ref{tokenizer} shows the comprehensive analyses on tokenizer from the perspective of compression.

Section \ref{details} shows the details and visualization of the training and inference processes.

Section \ref{compatibility} demonstrates the poor compatibility of VAR and continuous tokenizer.

Section \ref{mask} depicts the visual results of ablations on the mask mechanism, which elevates generation diversity.

Section \ref{scaling} presents the visual results as model size and inference step scaling up.

Section \ref{intermediate} shows the visual results at intermediate steps.

Section \ref{filter} shows the results of different low-pass filters.

Section \ref{prompts} shows the prompts of the text-to-image generaton for Figure 1 in main manuscript.

\section{Comprehensive Analyses on Tokenizer}  \label{tokenizer}
\noindent \textbf{Tokenizer: discrete or continuous.} Data compression and reconstruction are vital for image generation, determining the performance upper bound of generation. Given the discrete nature and the mature categorical cross-entropy loss of languages, a commonly adopted strategy for visual autoregressive models is to discretize the data with VQ. However, compared to the discrete human-created language, natural image space is continuous and infinite. Quantization, specifically in VQVAE, inevitably introduces significant information loss. 

For the tokenizer, the VQ operation induces significant information loss, making compression stage the bottleneck for better generation. Further, the autoregressive paradigm, i.e., "predicting next tokens based on previous ones", is independent of whether the values are discrete or continuous. The only difficulty that restricts the adoption of continuous-valued tokenizer is the lack of proper loss function to model the per-token probability distribution, which is easily done with cross-entropy for discrete-valued tokenizer. To this end, following the pioneering MAR \cite{li2024autoregressive}, we adopt the diffusion model as loss function. Specifically, the autoregressive method predicts a vector for each token, which then serves as conditioning for the denoising network.

\begin{figure*}[t]
    \begin{center}
	\includegraphics[width=0.95\linewidth]{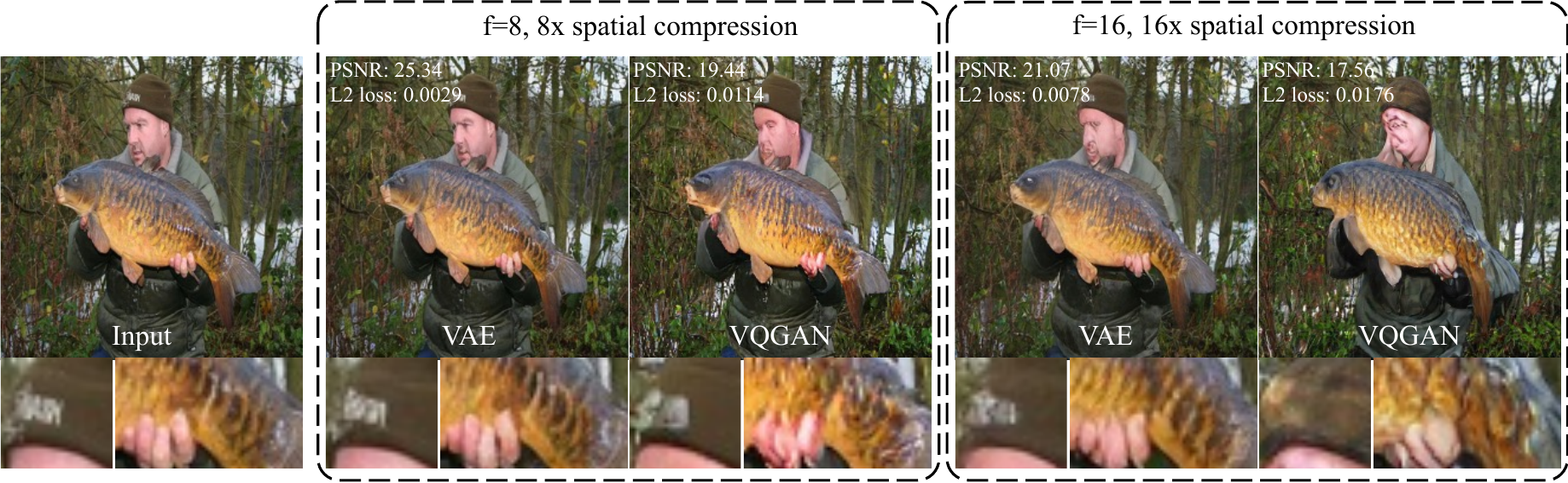}
    \end{center}
    \setlength{\abovecaptionskip}{-0.2cm}
    \setlength{\belowcaptionskip}{-0.2cm}
    \caption{Image reconstruction performance comparison between continuous and discrete tokenizers under different spatial compression ratios (f=8 and f=16). Constrained by their finite vocabulary codebooks, discrete tokenizers suffer from significant information loss, struggling to faithfully reconstruct images with intricate, high-frequency details such as human faces. Note that the reconstruction of continuous tokenizer at f=16 is still better than the discrete one at f=8, which is also consistent with the rate distortion theory.}
    \label{fig:reconstruction}
\end{figure*}

\begin{figure*}[ht]
    \begin{center}
	\includegraphics[width=0.98\linewidth]{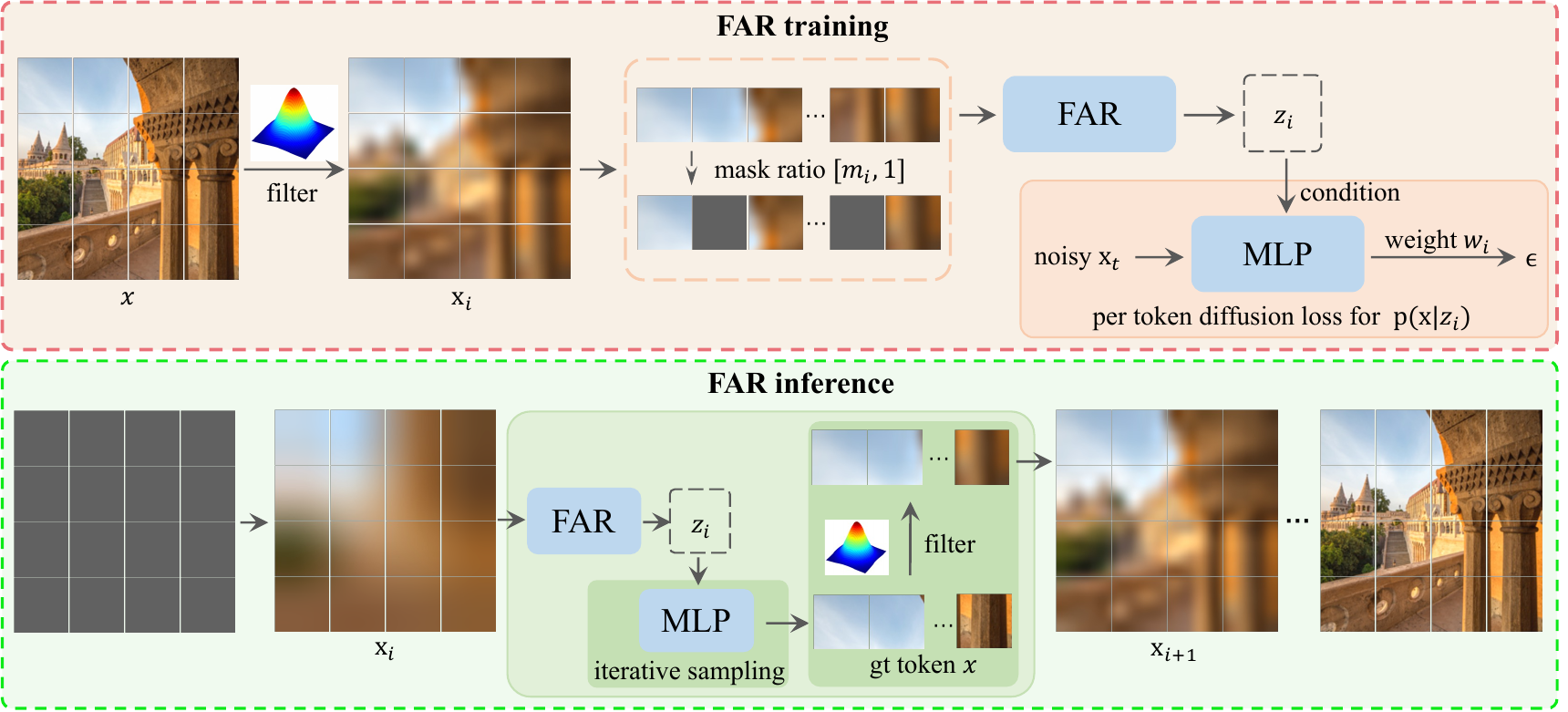}
    \end{center}
    \setlength{\abovecaptionskip}{-0.2cm}
    \setlength{\belowcaptionskip}{-0.2cm}
    \caption{\textbf{The visualization of the training and inference processes of FAR.} This flow chart demonstrates the details of FAR and its integration with continuous tokens.}
    \label{fig:sup_detail}
\end{figure*}

\noindent \textbf{Tokenizer: compression perspective.}
For images, latent space is crucial for generation for the purpose of reducing computational burden. Thus besides the data format (discrete or continuous), we further analyses the tokenizers, VQGAN or VAE, as compression models from two key aspects: 1) Theoretical compression performance; and 2) Reconstruction visual results. 

For measuring the theoretical compression performance, we adopt the information compression ratio (ICR) \cite{chen2024spark}. For discrete tokenizer, we take VQVAE as example, with downscaling factor $f$, codebook size $N$, input image's size of $H \times W$. We assume that the code follows a uniform distribution, so each code has $\log N$ bits information. For continuous tokenizer, we take VAE as example, with downscaling factor $f$, channel number $C$. Assume that the latent representation is fp32 tensor precision. The ICR of these two tokenizers are then as follows.
\begin{equation}
    \operatorname{ICR}(N, f)=\frac{(H / f) \times(W / f) \times \log N}{H \times W \times 3 \times \log 256}=\frac{\log N}{24 f^{2}}.
\end{equation}
\begin{equation}
    \operatorname{ICR}(C, f)=\frac{(H / f) \times(W / f)  \times C \times 32}{H \times W \times 3 \times \log 256}=\frac{32 C}{24 f^{2}}.
\end{equation}
Taking compression ration $f=16$ for example, $\operatorname{ICR}(N, f) = 0.23\% $ for discrete tokenizer with codebook size $N=16384$ and $\operatorname{ICR}(C, f) = 8.33\% $ for continuous tokenizer with channel number $C=16$. Further, to achieve same ICR under same $f$, we need to exponentially enlarge the codebook size $N$ from $16384$ to $2^{512} (1.34 \times 10^{154})$. 
Given that discrete tokenizers are inherently difficult to train \cite{yu2023magvit, mentzer2023finite}, it is thus prohibitively hard to train codebook at this scale.

In Figure \ref{fig:reconstruction}, we visualize the reconstructed images comparison between discrete and continuous tokenizers. Compared to continuous tokenizer, the discrete one has difficulty in both detail fidelity and semantic consistency. For instance, the character detail on the hat is poorly reconstructed by discrete tokenizer. For semantic, the face and fingers of the man as well as the fish scales fail to remain semantically identical by discrete tokenizer.

Based on the above analyses, the discrete tokenizer for images suffers from substantially more information loss than the continuous one, indicating that quantization, the shortcut stemming from mimicking languages autoregressive generation, may be a inferior solution for image data. Thus, apart from the commonly adopted VQ paradigm for image autoregressive models, it is quite necessary and promising to employ continuous tokenizer.

\section{Details and visualization of the training and inference processes}    \label{details}
In Figure \ref{fig:sup_detail}, we depict the flow chart of the training and inference processes of FAR, demonstrating the implementation details. \textit{For the training process}, FAR randomly selects a frequency level $i$ and filters the input image $x$ into the intermediate frequency level $x_i$. FAR then adopts the mask ratio $[r_i, 1]$ to the token sequence. The output $z_i$ of the FAR model is then conditioned on the diffusion MLP. The diffusion loss models the distribution of each token with the frequency-aware dynamic loss weight $w_i$. \textit{For the inference process}, we take the intermediate step $i$ as example. FAR takes the masked $x_i$ as input and outputs $z_i$. With $z_i$ as condition, the diffusion model samples the groundtruth token distribution $x$. Then, we can filter $x$ to get the next frequency level $x_{i+1}$.

\begin{figure*}[t]
    \begin{center}
	\includegraphics[width=0.98\linewidth]{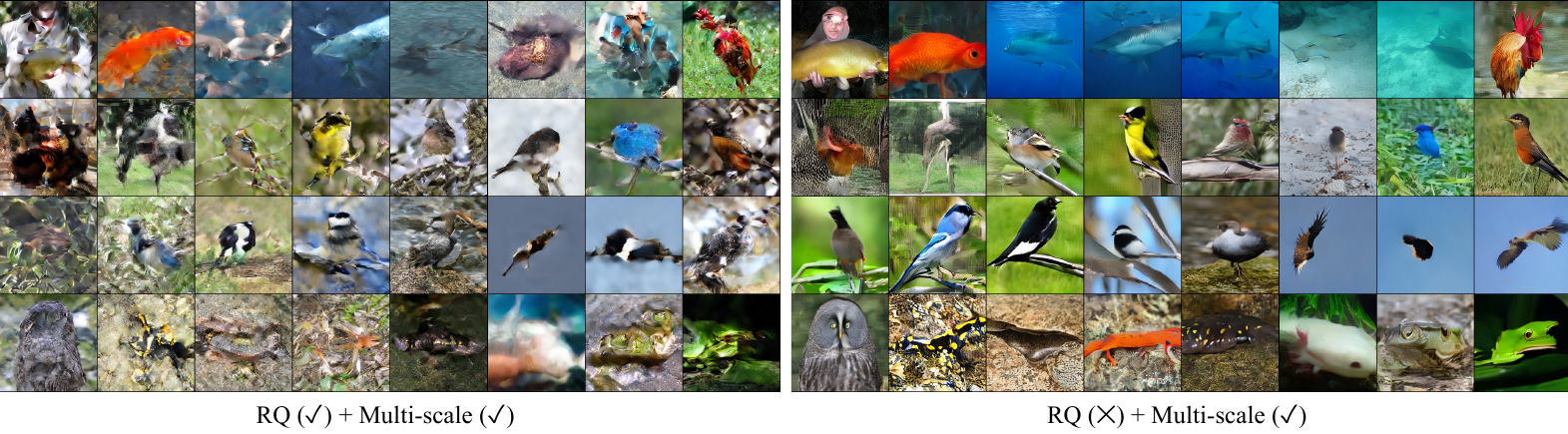}
    \end{center}
    \setlength{\abovecaptionskip}{-0.2cm}
    \setlength{\belowcaptionskip}{-0.2cm}
    \caption{\textbf{The visual results of combining VAR and continuous tokenizer.} These images correspond to the first 32 class labels. Consistent with our analyses, VAR paradigm demonstrates poor compatibility with the continuous tokenizer. The generation suffers from poor architectures and severe artifacts. Removing the residual quantization helps reducing the artifact, but still suffers from poor image quality.}
    \label{fig:sup_VAR}
\end{figure*}

\section{Compatibility of VAR and Continuous Tokenizer: Visual Results}    \label{compatibility}
In Figure \ref{fig:sup_VAR}, we depict the visual results of combining VAR and continuous tokenizer, corresponding to Table 4 of the main manuscript. The direct combination of VAR and continuous tokenizer demonstrates poor compatibility, generating images with obvious artifacts and poor architectures. On the right part, removing the Residual Quantization (RQ) successfully reduces the artifacts as expected. While, the generation is still inferior due to the challenging distribution modeling of diffusion loss in this case. These visual results exactly match our analyses in the manuscript.

\section{Visual Results of Ablations on the Mask Mechanism}    \label{mask}
\begin{figure*}[ht]
    \begin{center}
	\includegraphics[width=0.98\linewidth]{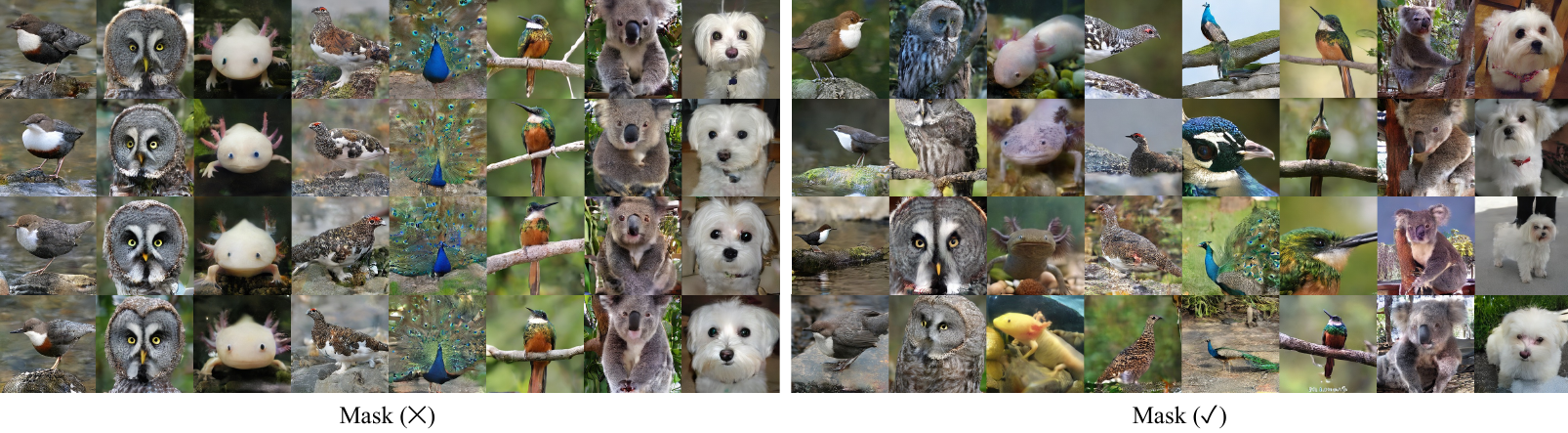}
    \end{center}
    \setlength{\abovecaptionskip}{-0.2cm}
    \setlength{\belowcaptionskip}{-0.2cm}
    \caption{\textbf{The visual results of ablations on the mask mechanism.} Each column corresponds to one class label. Our FAR can generate high-quality images without mask mechanism, while suffers from low diversity within each class. The mask strategy can effectively improves the generation diversity.}
    \label{fig:sup_mask}
\end{figure*}

In Figure \ref{fig:sup_mask}, we depict the visual results of the ablations on the mask mechanism, corresponding to Table 5 of the main manuscript. In the left part, our FAR can generate high-quality images after employing the diffusion loss distribution modeling simplification strategy. While, the generation diversity is limited. On the right part, we further adopt the mask mechanism. Mask introduces randomness, improving the generation diversity. 

\section{Visual Results as Model and Inference Step Scaling}        \label{scaling}
\begin{figure*}[ht]
    \begin{center}
	\includegraphics[width=0.98\linewidth]{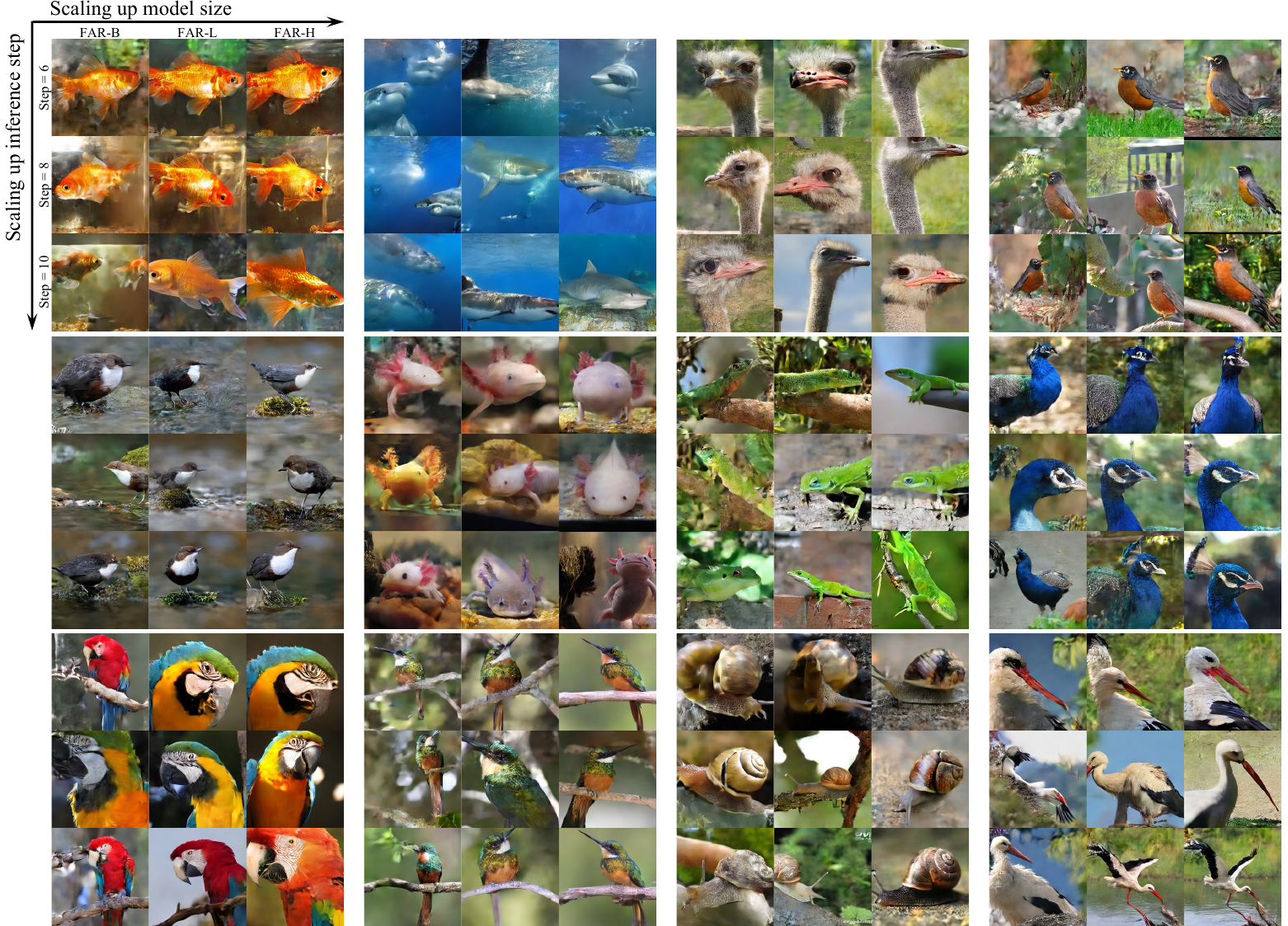}
    \end{center}
    \setlength{\abovecaptionskip}{-0.2cm}
    \setlength{\belowcaptionskip}{-0.2cm}
    \caption{\textbf{Visual Results as Model and Inference Step Scaling.} We depict the generation results when increasing the model size and inference step. Scaling up these two factors can consistently improve the generation performance.}
    \label{fig:sup_scaling}
\end{figure*}

In Figure \ref{fig:sup_scaling}, we verify the scaling capacity of FAR: including model size and inference step. Scaling up these two factors can consistently improve the generation performance.

\section{Visual Results at Intermediate Steps}          \label{intermediate}
\begin{figure*}[ht]
    \begin{center}
	\includegraphics[width=0.98\linewidth]{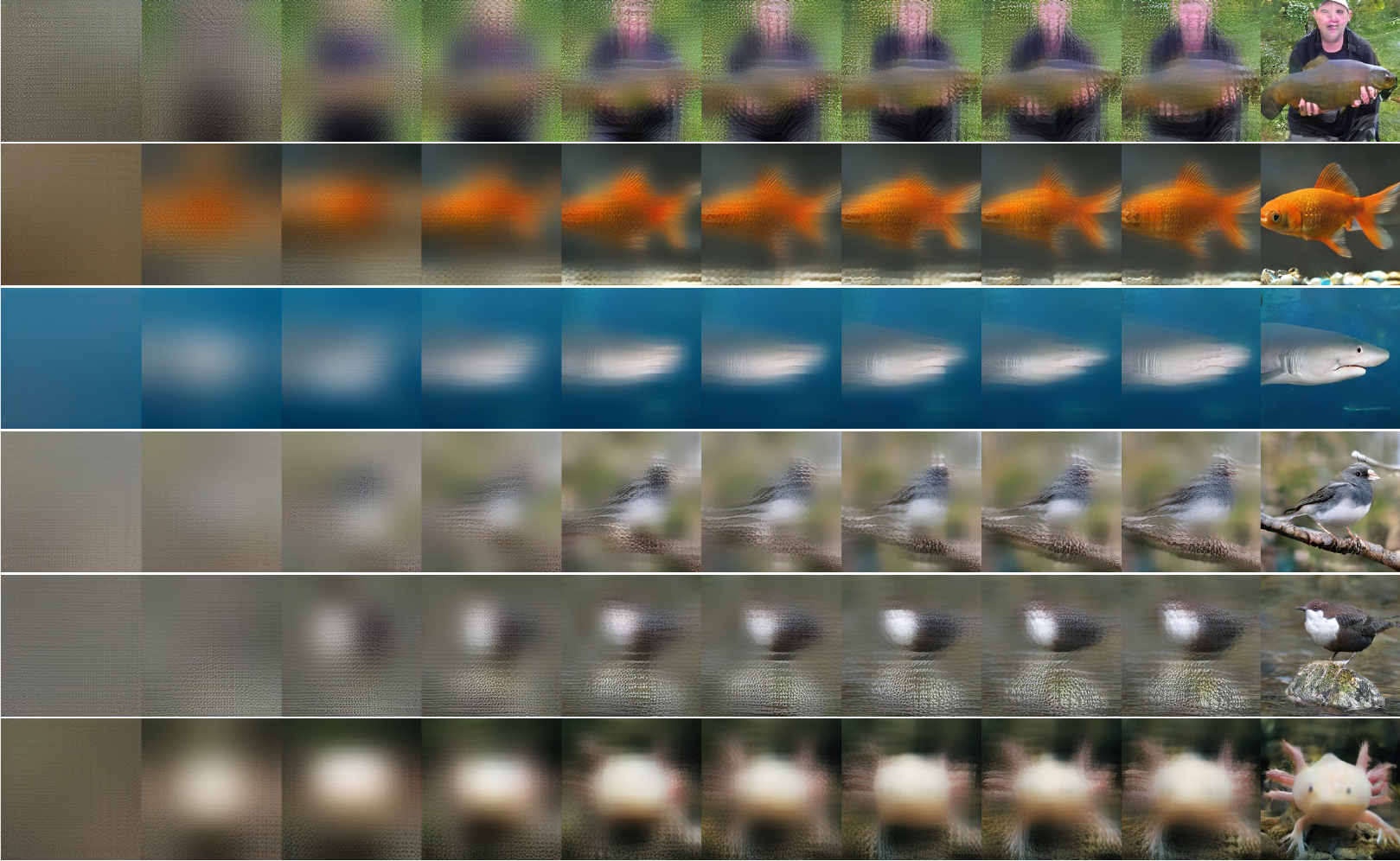}
    \end{center}
    \setlength{\abovecaptionskip}{-0.2cm}
    \setlength{\belowcaptionskip}{-0.2cm}
    \caption{\textbf{Visual results at intermediate steps.} The intermediate generation results (total step 10) autoregressively refine the details, aligning perfectly with our frequency progression design.}
    \label{fig:sup_interstep}
\end{figure*}

In Figure \ref{fig:sup_interstep}, we present the intermediate generation results along the autoregressive generation process. In the early steps, FAR generates the overall color and structural information, and then refines the details.

\section{Different low-pass filters}  \label{filter}
We explore two frequency filtering types: (a) first down-sample then up-sample in the spatial domain, (b) low-pass filter in the Fourier domain. We find that they yield similar performance, as shown in Table \ref{tab:Ablation}. Since different frequency filtering methods only slightly differ in the filter. Besides, our method processes the filtered image in the spatial domain, which further narrows the difference of different low-pass filters. We thus hold that the frequency filtering methods make small differences to the final performance. By default, we empirically adopt type (a) for simplicity.

\begin{table}[ht]
    \setlength{\tabcolsep}{12pt}
    \centering
    \vspace{-3mm}
    \caption{Results under different low-pass filters}
    \vspace{-3mm}
    \label{tab:Ablation}
    \begin{tabular}{c|cccc}
    \toprule
    Filters & FID$\downarrow$ & IS$\uparrow$ & Pre$\uparrow$ & Rec$\uparrow$ \\ 
    \midrule
    a  & 4.05 & 290.2 & 0.80 & 0.52   \\
    b  & 4.21 & 291.3 & 0.80 & 0.51   \\ 
    \bottomrule
    \end{tabular}
    \vspace{-3mm}
\end{table}

\section{Prompts for Figure 1 in Main Manuscript}      \label{prompts}
The following part presents the prompts for the text-to-image generation results in Figure 1 of the main manuscript: 
\begin{itemize}
    \item A mountain village built into the cliffs of a canyon, where bridges connect houses carved into rock, and waterfalls flow down into the valley below.
    \item An otherworldly forest of giant glowing mushrooms under a vibrant night sky filled with distant planets and stars, creating a dreamlike, cosmic landscape.
    \item A close-up photo of a bright red rose, petals scattered with some water droplets, crystal clear.
    \item A photo of a palm tree on water. 
    \item A bird made of crystal.
    \item A tranquil scene of a Japanese garden with a koi pond, painted in delicate brushstrokes and a harmonious blend of warm and cool colors.
    \item Paper artwork, layered paper, colorful Chinese dragon surrounded by clouds.
    \item A still life of a vase overflowing with vibrant flowers, painted in bold colors and textured brushstrokes, reminiscent of van Gogh's iconic style.
    \item A peaceful village nestled at the foot of towering mountains in a tranquil East Asian watercolor scene. 
    \item An enchanted garden where every plant glows softly, and creatures made of light and shadow flit between the trees, with a waterfall flowing in the background.
    \item A lion teacher wears a suit in the forest.
    \item A cloud dragon flying over mountains, its body swirling with the wind.
\end{itemize}

\end{document}